\documentclass{isprs} 
\usepackage{subfigure}
\usepackage{soul}
\usepackage{setspace}
\usepackage{geometry} 
\usepackage{epstopdf}
\usepackage[labelsep=period]{caption}  
\usepackage[british]{babel} 
\usepackage[hang]{footmisc}
\usepackage{booktabs}
\usepackage{multirow}
\usepackage{makecell}
\usepackage{url}
\usepackage{siunitx}
\usepackage{xurl}
\usepackage{xcolor}
\usepackage{amssymb}
\usepackage{makecell}
\usepackage{hyperref}
\usepackage{amsmath}
\usepackage{stfloats}

\usepackage{graphicx}
\usepackage{comment}

\hypersetup{
    colorlinks=true,    
    citecolor=[rgb]{0,0,.6},
    urlcolor =[rgb]{0,0,.6},
    linkcolor = blue
    }

\begin{document}

\title{PC2Model: ISPRS benchmark on 3D point cloud to model registration
}
\sloppy

\author{
 Mehdi Maboudi\textsuperscript{1}, Said Harb\textsuperscript{1}, Jackson Ferrao\textsuperscript{1}, Kourosh Khoshelham\textsuperscript{2}, Yelda Turkan\textsuperscript{3}, Karam Mawas\textsuperscript{1}}

\address{
\textsuperscript{1 }Institute of Geodesy and Photogrammetry, Technische Universit\"at Braunschweig, Germany \\{(m.maboudi, s.harb, j.ferrao, k.mawas)}@tu-braunschweig.de\\\textsuperscript{2 }Department of Infrastructure Engineering, University of Melbourne, Australia.
k.khoshelham@unimelb.edu.au\\
\textsuperscript{3 }Civil \& Construction Engineering,
Oregon State University, USA.
yelda.turkan@oregonstate.edu
}
\date{}

\abstract{
Point cloud registration involves aligning one point cloud with another or with a three-dimensional (3D) model, enabling the integration of multimodal data into a unified representation. This is essential in applications such as construction monitoring, autonomous driving, robotics, and virtual or augmented reality (VR/AR). With the increasing accessibility of point cloud acquisition technologies, such as Light Detection and Ranging (LiDAR) and structured light scanning, along with recent advances in deep learning, the research focus has increasingly shifted towards downstream tasks, particularly point cloud-to-model (PC2Model) registration. While data-driven methods aim to automate this process,  they struggle with sparsity, noise, clutter, and occlusions in real-world scans, which limit their performance. 
To address these challenges, this paper introduces the PC2Model benchmark, a publicly available dataset designed to support the training and evaluation of both classical and data-driven methods. Developed under the leadership of ICWG II/Ib, the PC2Model benchmark adopts a hybrid design that combines simulated point clouds with, in some cases, real-world scans and their corresponding 3D models. Simulated data provide precise ground truth and controlled conditions, while real-world data introduce sensor and environmental artefacts. This design supports robust training and evaluation across domains and enables the systematic analysis of model transferability from simulated to real-world scenarios. The dataset is publicly accessible at: \href{https://doi.org/10.5281/zenodo.17581812}{https://zenodo.org/records/17581812}.
}


\maketitle

\begin{figure*}[b]
\centering
\includegraphics[width=1\textwidth]{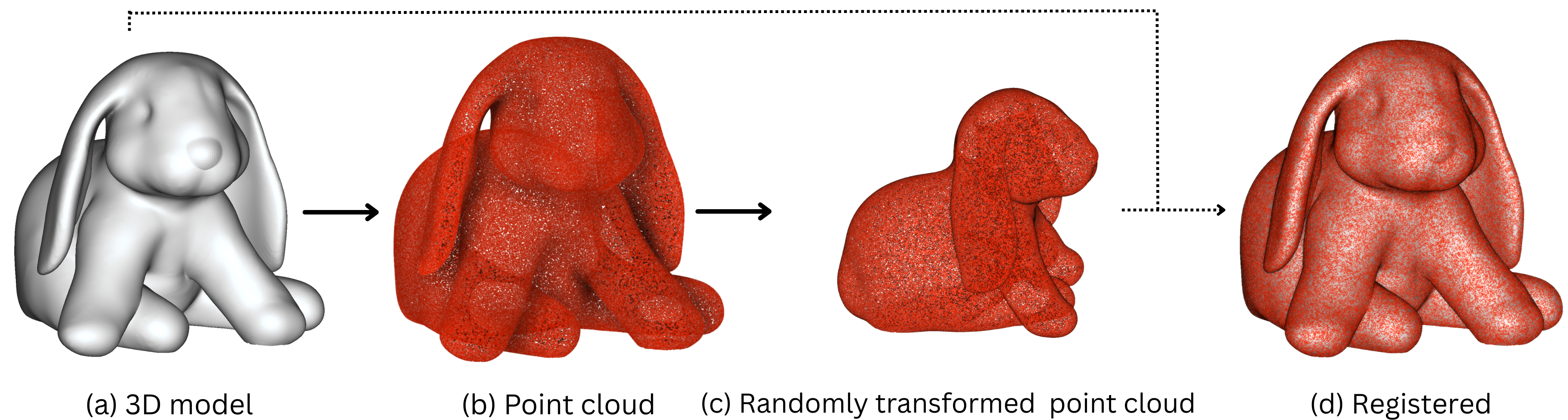}
\caption{Point cloud simulation from 3D model and its co-registration. a) 3D model, b) point cloud sampled from 3D model, c) point cloud after arbitrary 7-parameter transformation, and d) transformed point cloud registered to the 3D model.}
\label{fig:pc2model_process}
\end{figure*}


\section{Introduction}

The proliferation of three-dimensional (3D) sensing technologies has fundamentally transformed how physical environments are captured and analysed. Devices such as Light Detection and Ranging (LiDAR) scanners, photogrammetric systems, and structured light sensors now produce dense geometric representations of real-world scenes at scale, and their growing affordability has made such data commonplace across disciplines~\cite{maboudi_25}. In fields ranging from construction site monitoring and autonomous driving to robotics, digital 3D printing, and virtual and augmented reality (VR/AR), there is a pressing need to interpret these representations in relation to prior geometric knowledge, most often in the form of 3D models~\cite{huang_24,mora_21}. This has driven continued interest in point cloud registration: the task of determining the spatial transformation that best aligns a captured point cloud with a reference, whether that reference is another point cloud or a 3D model. While point cloud-to-point cloud registration has been studied extensively and is supported by a variety of established benchmarks, the analogous problem of point cloud-to-model registration remains relatively underexplored. Existing datasets do not generally accommodate this setting, limiting the ability to train and evaluate data-driven approaches. To address this, we introduce the ISPRS benchmark on 3D point cloud-to-model registration, PC2Model. It features 3D models and their transformed scans for alignment (co-registration). Figure~\ref{fig:pc2model_process} illustrates the co-registration process schematically.

Algorithms for point cloud registration encompass a broad range of techniques, from established methods such as iterative closest point (ICP) to deep learning-based approaches~\cite{zhao_22}. In practice, point clouds obtained from real-world scans present challenges such as sparsity, clutter, noise, mixed pixels, and occlusions, which reduce the performance of point cloud registration algorithms. The PC2Model benchmark includes these artefacts in a controlled and systematically reduced manner, providing a dataset that closely approximates real-world conditions. Developed under the leadership of ICWG~II/I~b, PC2Model provides a publicly accessible dataset for 3D point cloud-to-model registration, encompassing 137 samples. A key feature of the dataset is its hybrid design, combining simulated point clouds, and, in some cases, real-world scans with their corresponding 3D models. The simulated point clouds offer precise ground truth alignments and controlled acquisition conditions, whereas the real-world data introduce sensor- and environmentally induced artefacts such as beam divergence, noise, and measurement inaccuracies. This hybrid configuration facilitates the training and evaluation of data-driven registration methods across both domains and enables systematic investigation of the transferability of models trained on simulated data to real-world scenarios.

The rest of this paper is organised as follows:
In section \ref{Related Work}, we review point cloud-to-point cloud and point cloud-to-mesh co-registration methods, and discuss existing datasets relevant to point cloud-to-model co-registration.
Section \ref{Dataset} describes the PC2Model dataset and includes the categories of objects and details of the point cloud simulation of the 3D models.  Section \ref{Simulation Env.} describes the simulation environment and the implemented add-on. Section \ref{Evaluation} then presents and discusses evaluation strategies and metrics. Finally, section \ref{Conclusion} summarises the paper and outlines directions for future work.

\section{Related Work}\label{Related Work}

The registration of point clouds to 3D models is a fundamental problem in computer vision, robotics, and photogrammetry. Over the years, a wide range of classical and learning-based methods have been developed, each with its own strengths and limitations. This section reviews representative approaches, spanning conventional techniques to modern learning-based methods. 

\textbf{Conventional methods:} The Iterative Closest Point (ICP) algorithm~\cite{besl_1992} is one of the most widely used methods for rigid registration. It iteratively establishes correspondences between a source and a target (point cloud or model), and computes the rigid transformation that minimises the distances between matched points. ICP performs well when a good initial coarse alignment is available; however, it is sensitive to local minima, symmetry ambiguities, or insufficient geometric features~\cite{Mawas_2026}. To address these limitations of naive ICP, Go-ICP~\cite{yang_2016} employs a branch-and-bound strategy to find the optimality of the rigid transformation. The Normal Distributions Transform (NDT)~\cite{biber_2003} adopts a probabilistic approach by representing the target cloud as a set of Gaussian distributions over discretised voxels and aligning the source points by maximising the likelihood. Unlike ICP, NDT does not rely on explicit point-to-point correspondences, improving robustness to variations in point density, noise, and partial overlap. Fast Global Registration (FGR)~\cite{zhou_2016} provides an efficient alternative to sampling-based approaches by directly optimising a robust objective over a dense set of candidate correspondences derived from Fast Point Feature Histograms (FPFH). Using a scaled Geman-McClure estimator with graduated non-convexity, FGR suppresses outliers without requiring iterative closest-point queries, achieving accuracy comparable to well-initialised local refinement methods such as ICP, while offering significant computational gains.

\textbf{Learning-based methods:} To tackle the co-registration problem, learning-based methods have been proposed. Deep Closest Point (DCP)~\cite{wang_2019} learns pointwise features from source and target clouds and predicts the rigid transformation via a differentiable SVD (singular value decomposition) layer. PointNetLK~\cite{sarode_2019} combines the PointNet architecture with the classic Lucas–Kanade alignment method, extracting global features, and iteratively updating the transformation to minimise feature differences. These methods demonstrate how feature learning can improve robustness and generalisation across a wide range of point cloud shapes and densities. In summary, classic algorithms such as ICP, Go-ICP, and NDT remain foundational for rigid registration, while learning-based approaches such as DCP and PointNetLK show promise in handling complex and diverse point cloud scenarios. 

\textbf{Point cloud-to-model registration:} In contrast to the prior-mentioned studies that use point clouds as reference data, using a mesh as a reference can improve robustness by explicitly enforcing visibility (line-of-sight) constraints during correspondence selection. In this framework, synthetic range images are rendered from the meshes and integrated into a Monte Carlo Localisation (MCL) pipeline~\cite{Chen_2021}. In contrast, ICP-like methods have incorporated ray-casting correspondences (RCC) to register sensor data with a mesh model~\cite{Vizzo_2021}. RCC is introduced to reduce incorrect matches in challenging configurations such as thin structures and self-occlusions, but the presented implementation did not achieve real-time performance. RCC efficiency is substantially improved in~\cite{Mock_2024} by leveraging heterogeneous parallel hardware, while estimating the rigid transformation using a standard SVD-based approach. A geometry-based approach by~\cite{zhao_22} addresses cross-form registration between LiDAR point clouds, BIM models, and HoloLens meshes by extracting planar surfaces and matching them across data types, achieving this without manual initialisation.

\textbf{Existing datasets}  related to point cloud to 3D model alignment mainly originate from adjacent tasks such as object pose estimation or multimodal registration. The LINEMOD~\cite{Hinterstoisser_2013} and TUD-L~\cite{Hodan_2018} datasets, provided within the BOP benchmark~\cite{Hodan_2018}, pair object-centric point clouds derived from depth images with corresponding CAD models and ground-truth 6D poses. These datasets primarily focus on instance-level object pose estimation in cluttered scenes and cover only a small number of object categories. LINEMOD comprises 15 objects and TUD-L just 3, making them limited in scope for benchmarking registration methods that must generalise across diverse geometries. The 3DPCD-CT dataset~\cite{saiti_2022} addresses the registration of point clouds to surface representations by providing paired point clouds and mesh-like models with known rigid transformations. While this dataset explicitly targets point cloud to model alignment, it is limited in scale and variety and focuses on a specific application domain. In contrast, PC2Model is designed as a dedicated benchmark for point cloud-to-3D-model registration at a broader scale, comprising 137 object instances that span a wide variety of categories, shapes, and scanning conditions, including varying point densities, viewpoints, and realistic scanning artefacts. Unlike existing datasets, PC2Model emphasises generalisation across object classes and challenging initial misalignments, making it well-suited for evaluating both classical and learning-based registration algorithms under realistic conditions.

\begin{table}[t]
  \centering
  \resizebox{\columnwidth}{!}{%
    \begin{tabular}{l l c}
      \toprule
      Source      & Category          & Number of samples \\
      \midrule

      \multirow{6}{*}{\makecell[c]{Simulated\\from\\3D model}}
        & \multicolumn{1}{|l}{Mechanical objects} & 25 \\
        & \multicolumn{1}{|l}{Furniture} & 25 \\
        & \multicolumn{1}{|l}{Home décor} & 25 \\
        & \multicolumn{1}{|l}{Houses} & 25 \\
        & \multicolumn{1}{|l}{Vehicles} & 25 \\
        & \multicolumn{1}{|l}{Indoor spaces} & 6  \\

      \cline{1-1}
      \multicolumn{1}{l}{\raisebox{-3pt}{Real}}
        & \multicolumn{1}{|l}{\raisebox{-2pt}{Indoor spaces}    }                   & 6  \\

      \midrule
      &                                    & \textbf{Total}: 137 \\
      \bottomrule
    \end{tabular}%
  }
    \caption{Categories and number of samples. The Indoor spaces (real) are from \protect\cite{Khoshelham_2020}. }
    \label{tab:dataset_categories}  
    
\end{table}

\begin{table}[b]
    \centering
    \setlength{\tabcolsep}{3pt}
    \small
    \resizebox{\columnwidth}{!}{%
    \begin{tabular}{l l
                    S[table-format=2.0]
                    S[table-format=8.0]
                    S[table-format=5.0]}
        \toprule
        Source &
        Category &
        {\makecell{Avg. \\nr. of scans}} &
        {\makecell{Avg. \\nr. of points}} &
        {\makecell{Avg. median\\point density\\($r = 0.5$\,m)}} \\
        \midrule
        \multicolumn{1}{c|}{\multirow{6}{*}{\makecell[c]{Simulated\\from\\3D model}}}
            & Mechanical objects    & 6   & 191961    & 5495  \\
        \multicolumn{1}{c|}{}
            & Furniture             & 6   & 221275    & 67    \\
        \multicolumn{1}{c|}{}
            & Home décor            & 5   & 244270    & 4881  \\
        \multicolumn{1}{c|}{}
            & Houses                & 19  & 24823458  & 47542 \\
        \multicolumn{1}{c|}{}
            & Vehicles              & 4   & 277288    & 19228 \\
        \multicolumn{1}{c|}{}
            & Indoor spaces         & 13  & 21009545  & 43872 \\
        \cline{1-1}
        \multicolumn{1}{c|}{Real}
            & Indoor spaces  & {N/A} & 19507569 & 30127 \\
        \bottomrule
    \end{tabular}%
    }
    \caption{Dataset statistics. All scans were performed with simulated terrestrial laser scanners, except for indoor spaces (Real).}
    \label{tab:statistics}
\end{table}

\section{Dataset}\label{Dataset}
The PC2Model benchmark dataset comprises simulated and real-world point clouds divided into seven categories. The six simulated categories and one real-world category are shown in Table~\ref{tab:dataset_categories}, along with the number of samples. The real category contains scans acquired in real-world indoor environments with a laser scanner. Each sample includes a 3D model, the corresponding point cloud generated by either a simulated or a real laser scanner, and a rigid transformation applied to the scan, as well as the associated transformation matrices.

The dataset statistics are presented in Table \ref{tab:statistics}. Houses and indoor scenes represent the largest objects with the most complex geometry; therefore, they typically have a higher number of scanners, a larger number of points, and a higher average median point density. The object categories (mechanical objects, furniture, vehicles, and home décor) generally require fewer scanners to cover the entire object, as the objects are smaller and exhibit fewer occlusions than houses with multiple rooms and levels. Point densities for houses and indoor scenes are significantly higher than for object categories, where objects are scanned in open space. The lowest point densities are observed in furniture scans. Point density was computed by forming a sphere around each point in the point cloud and counting the number of points within the sphere. This count was then divided by the volume of the sphere, and the median of these densities is calculated. For houses, vehicles and indoor scenes, the radius can be interpreted as meters; however, this does not necessarily apply to the other categories.

\begin{figure}[t]
\centering
\includegraphics[width=1.0\columnwidth]{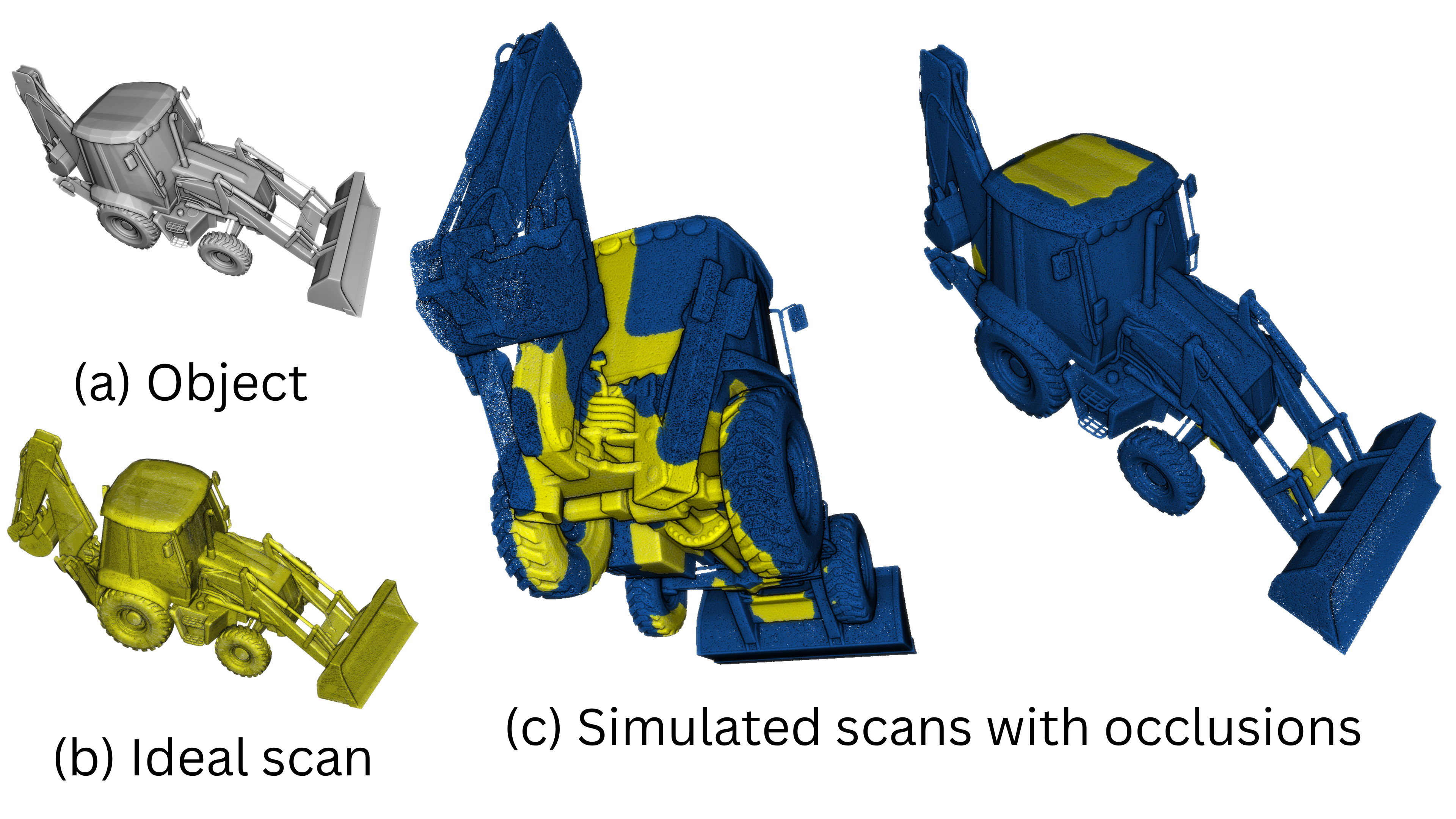}
\caption{Occluded regions (yellow) and scanned regions (blue) compared to an ideal scan, shown from an inverted viewpoint.}
\label{fig:occulusion}
\end{figure}

\begin{figure}[b]
\centering
\includegraphics[width=0.7\columnwidth]{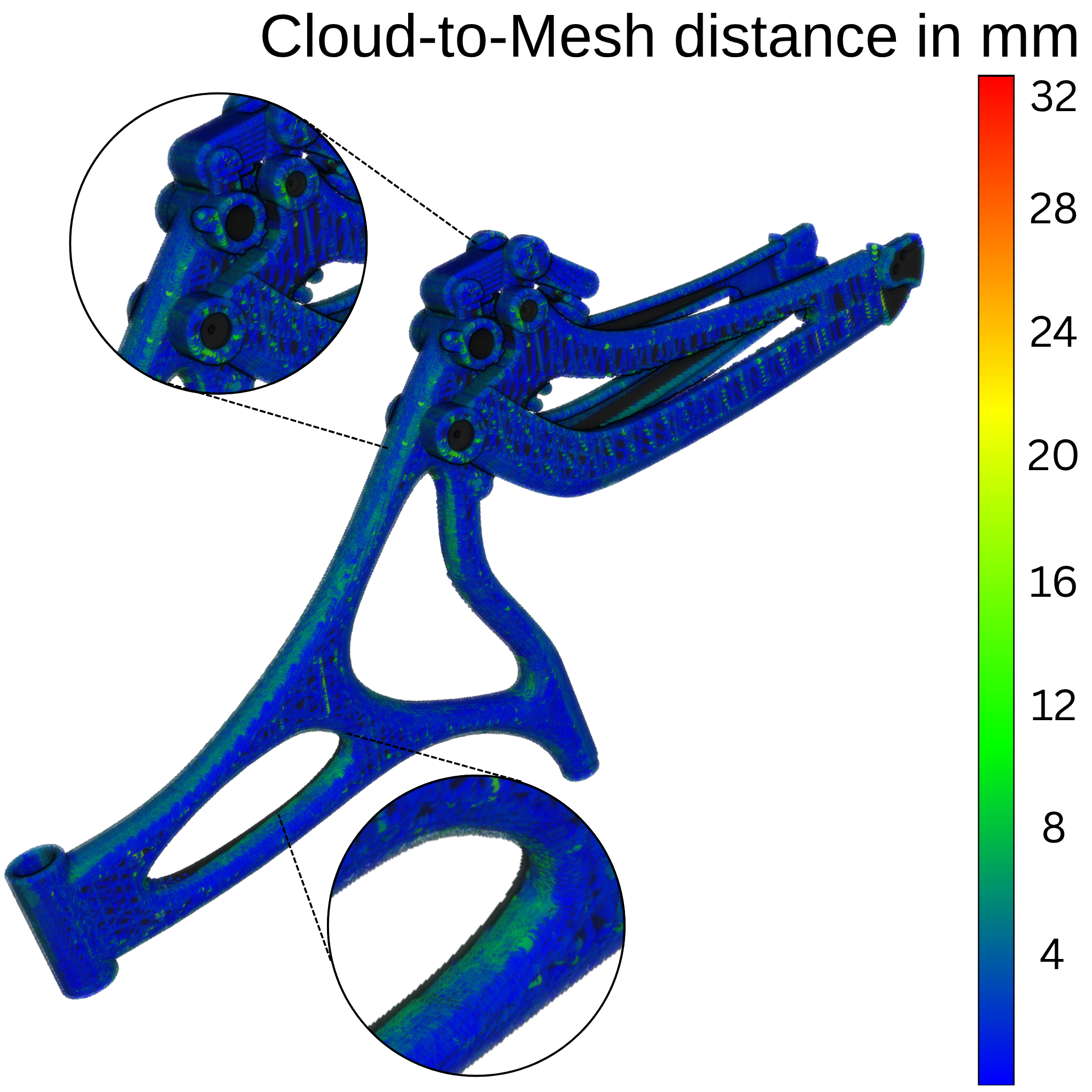}
\caption{Surface deviation from the ground-truth model.}
\label{fig:noise}
\end{figure}

\begin{figure*}[h]
\centering
\includegraphics[width=0.48\textwidth]{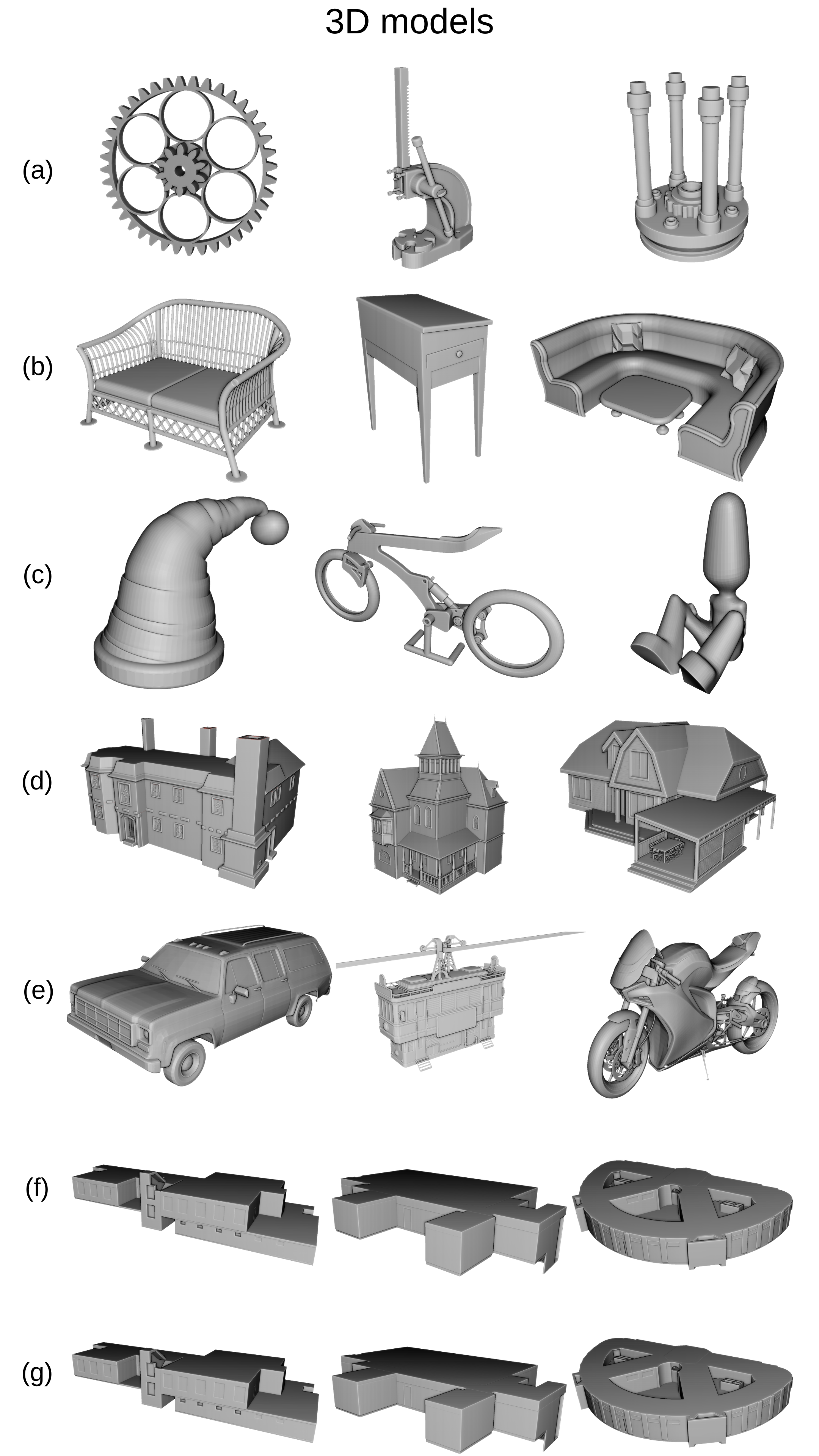}
\hfill
\includegraphics[width=0.48\textwidth]{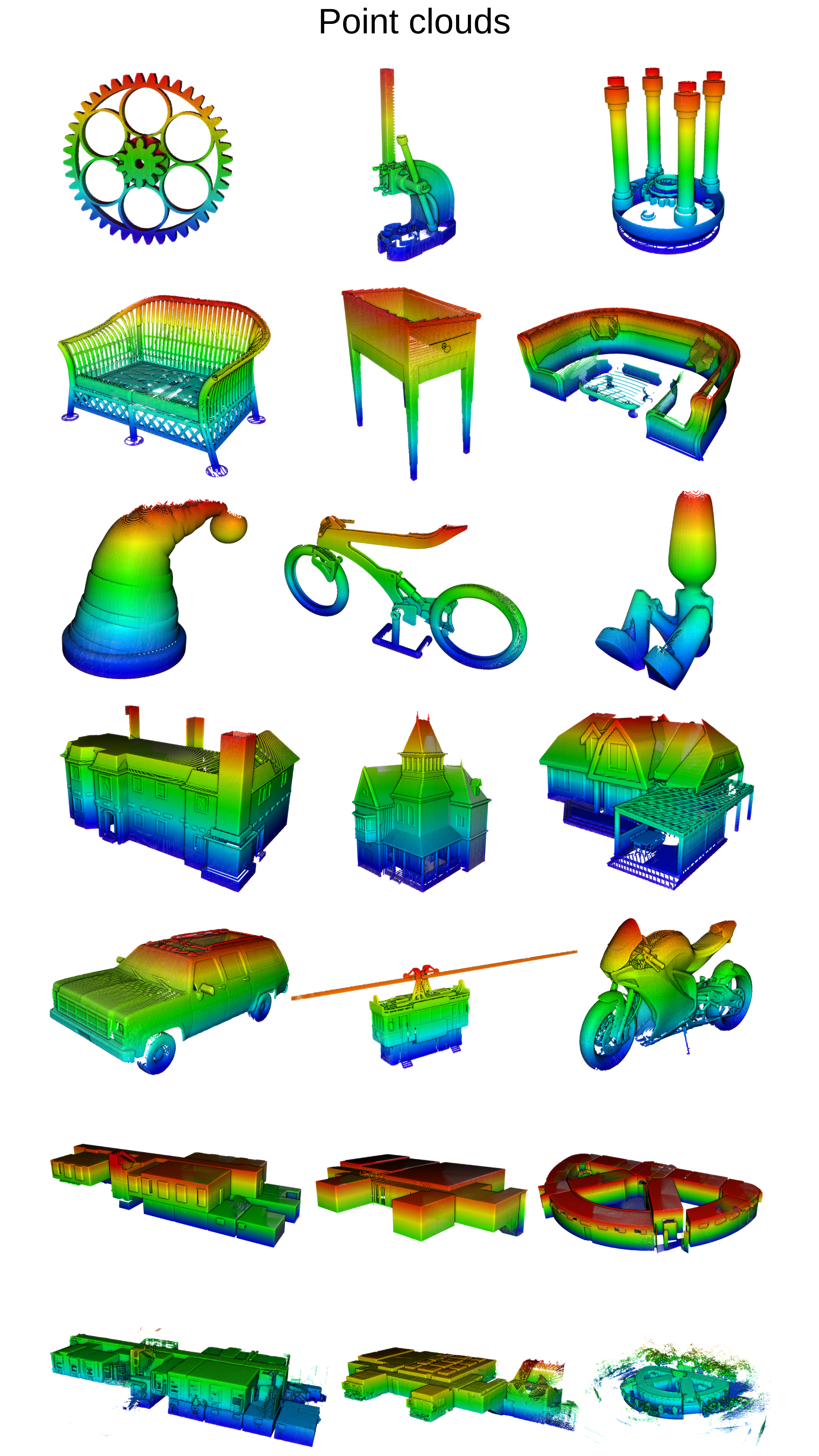}

\caption{Representative dataset samples spanning seven categories: (a) mechanical objects, (b)furniture, (c) home décor, (d) houses, (e) vehicles,  (f) indoor spaces, and (g) indoor spaces (real). Each sample is shown as a point cloud (right) alongside its corresponding 3D reference model (left).}
\label{fig:pc_and_obj_samples}
\end{figure*}

Depending on the model category, different scanning strategies were used. For mechanical objects, furniture, vehicles, and home décor, scanners were placed on the ground plane surrounding the object. As a result, the upper parts of these objects may not always be fully scanned. In addition, vehicles, houses, and indoor scenes were scaled to their approximate real-world dimension. The rest of the object categories were intentionally not scaled to their real-world sizes to introduce additional size variability into the dataset. Some houses are furnished, while others contain empty rooms; however, the scanning strategy remained the same.
Scanners were placed at approximately tripod height (\SI{1.5} {\meter} to \SI{2}{\meter}) above ground level in each room, where additional areas, such as the roof 
spaces, were accessible, for example, through connecting doors; further scanners were placed accordingly. Furthermore, scanners were positioned outside the houses at ground level to capture the exterior walls. For indoor scenes, scanners were also placed in each room. In contrast to houses, no scanners were positioned outside indoor scenes; therefore, the exterior walls were not scanned. The real-world indoor scans are from the ISPRS benchmark for indoor modeling~\cite{Khoshelham_2020} and feature six different buildings with various complexities. In Figure \ref{fig:pc_and_obj_samples}, a visual overview of samples of each category is depicted. 

\begin{figure}[t]
\centering
\includegraphics[width=0.7\columnwidth]{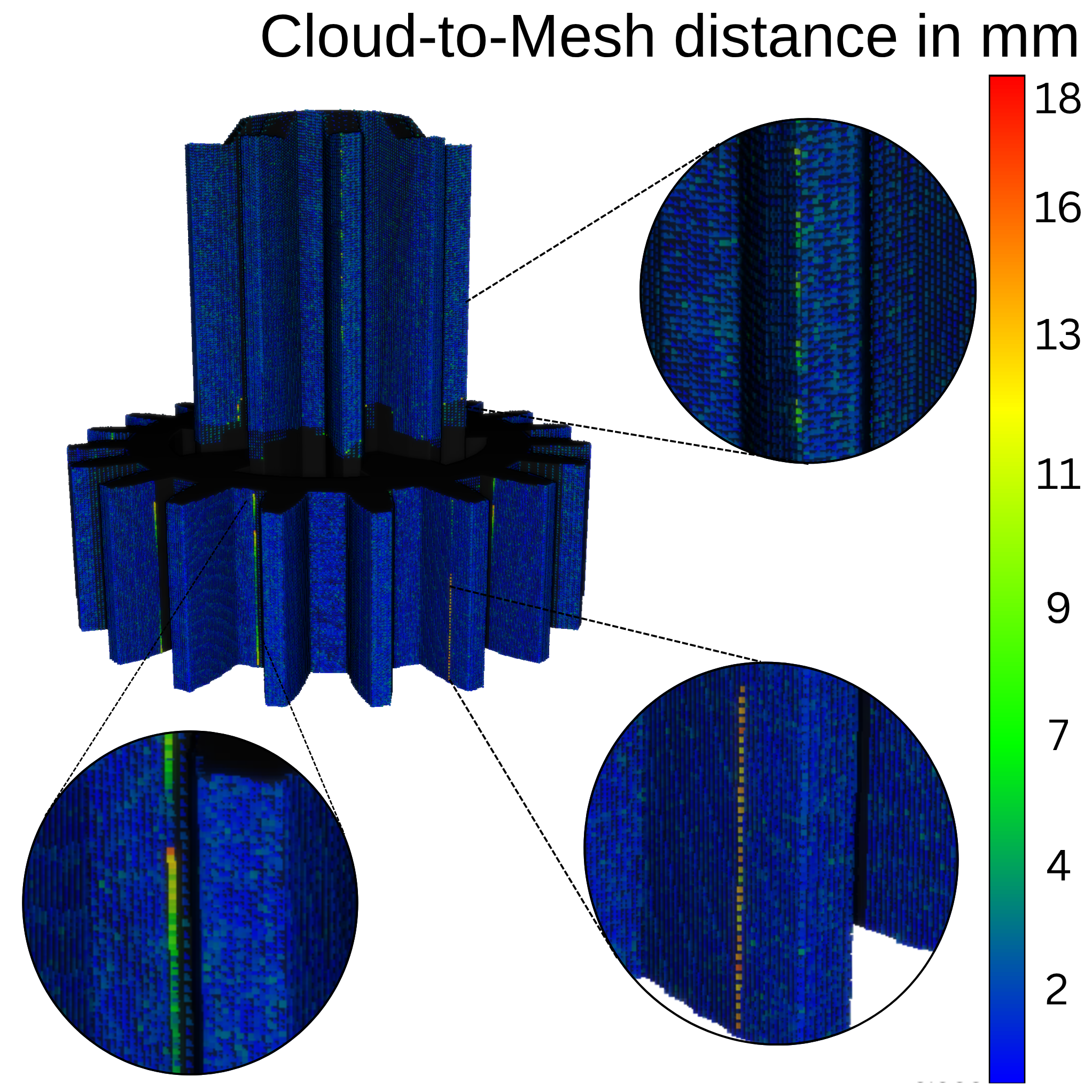}
\caption{Mixed-pixel artefacts along object edges, caused by partial beam hits on multiple surfaces.}
\label{fig:mixed_pixel}
\end{figure}

\begin{figure}[b]
\centering
\includegraphics[width=0.7\columnwidth]{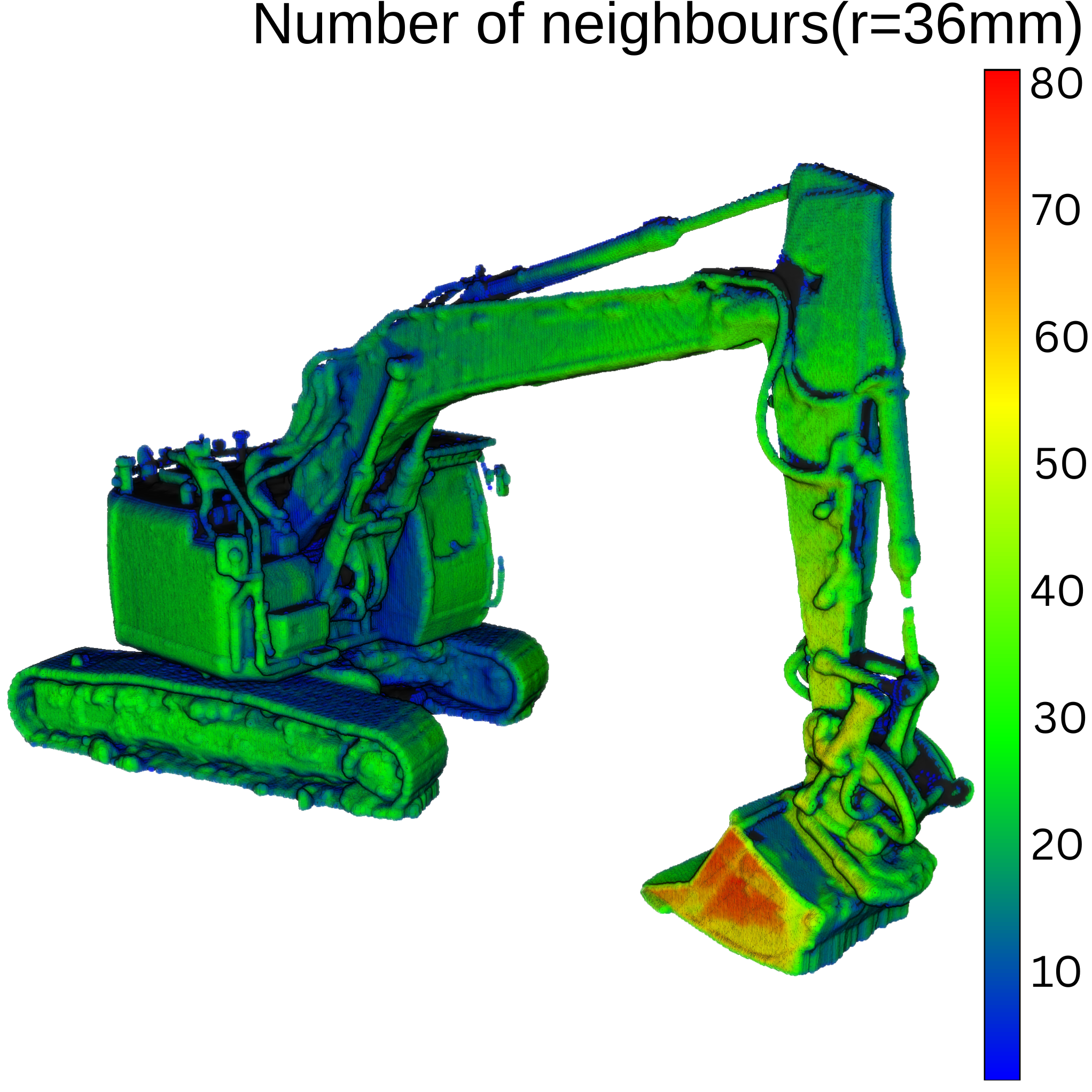}
\caption{Point density variation across the surface; warmer colors indicate higher local density.}
\label{fig:var_in_density}
\end{figure}

The PC2Model benchmark includes common real-world scanning artefacts in a controlled and systematically reduced form, approximating realistic scanning conditions without eliminating these effects entirely. Mixed pixels arise along object edges where the laser beam partially hits multiple surfaces, producing spurious points, as shown in Figure~\ref{fig:mixed_pixel}. By inspecting Cloud-to-Mesh distance, depicted in Figure~\ref{fig:noise}, noise manifests as surface deviations from the ground-truth model, visible as localised point displacements. Occlusions result in missing regions where surfaces are not visible from the scanner's perspective (Figure~\ref{fig:occulusion}). Finally, density variations occur due to differences in scan angle and distance, causing uneven point distributions across the object surface (Figure~\ref{fig:var_in_density}).

All provided point clouds underwent a rigid-body transformation, which serves as the ground truth for point cloud-to-model registration algorithms. The transformed point clouds, the original three-dimensional models, and the corresponding homogeneous transformation matrices are provided in the dataset. The transformations were applied as follows. For each object, a point cloud scan is simulated and then rotated around the $x$-, $y$-, and $z$-axes using independent random angles between \SI{0}{\degree} and \SI{360}{\degree}. The point cloud is then translated. Due to the fact that not every model has a true scale, the translation is conducted relatively. As a result, a maximum translation factor $f$ with $5>=f>=-5$ is defined. For each spatial direction ($x$, $y$, $z$), a random value between the negative and positive maximum translation factor was selected and multiplied by the extent of the point cloud in that direction. This ensures that the applied translations are meaningful regardless of the scale of the point cloud. Furthermore, the point clouds were scaled with a chance of 50\% by a scaling factor between 0.5 and 1.5.

The sources of the 3D models for the simulated point clouds range from publicly available datasets, including the ABC dataset~\cite{ABC}, ModelNet~\cite{ModelNet10}, Fusion360~\cite{Fusion}, and Thingi10k~\cite{Thingi10K}, to user-contributed, open-licensed 3D models hosted on the Sketchfab\footnote{\url{ https://sketchfab.com}} platform. In addition, for real-world scans, the ISPRS indoor modelling dataset~\cite{Khoshelham_2020} was included in the benchmark. All models were carefully curated to meet the diversity and quality requirements of the PC2Model dataset.

\section{Simulation software and Blender add-on}\label{Simulation Env.}

\begin{table}[t]
    \centering
    \begin{tabular}{lS[table-format=7.3]}
        \toprule
        \textbf{Parameter} & \textbf{Value} \\
        \midrule
        Accuracy & \SI{1.2}{\milli\meter} \\
        Beam divergence & \SI{0.23}{\milli\radian} \\
        Maximum rotation speed & \SI{18}{\degree\per\second} \\
    Pulse frequencies & {\makecell[c]{\SI{125000}{\hertz} \\[2pt]
                                     \SI{250000}{\hertz} \\[2pt]
                                     \SI{500000}{\hertz} \\[2pt]
                                     \SI{1000000}{\hertz}}} \\
        Pulse length & \SI{0.75}{\nano\second} \\
        Minimum scan range & \SI{0.4}{\meter} \\
        Maximum vertical scan angle & \SI{145}{\degree} \\
        Wavelength & \SI{1550}{\nano\meter} \\
        Minimum scan frequency & \SI{4}{\hertz} \\
        Maximum scan frequency & \SI{100}{\hertz} \\
        \bottomrule
    \end{tabular}
    \caption{Specifications of the Leica ScanStation P40 laser scanner}
    \label{tab:scanner}
\end{table}

To simulate the scanning of the 3D models, Helios++~\cite{heliosPlusPlus} was selected from several available simulation tools. Helios++ recreates laser-scanning environments with high fidelity, ensuring that the generated point clouds resemble real-world scans. The software is open source and supports the import of custom 3D models in Wavefront Object (OBJ) format. In
combination with configurable simulation parameters and transparent file handling, these features enable realistic scan behaviour at a manageable computational cost. Helios++ supports simulations for terrestrial, airborne, and mobile laser scanners. In this project, terrestrial scanning was used. Among the available scanning methods, polygonal mirror deflection was selected. A key feature of Helios++ is its waveform-based laser beam simulation. Instead of modelling beams as straight lines, beam footprints are represented with a finite spatial extent, and
reflected energy is integrated to derive surface characteristics. Beam divergence is modelled using multiple sub-rays sampled around a central ray, each of which is simulated individually, enabling
control over parameters that affect scanning accuracy. Ray tracing is used to compute intersections between rays and scene geometry, and additional noise sources, including range measurement noise and sensor position uncertainty, are
incorporated to generate more realistic point clouds. As Helios++ provides a high-fidelity laser scanner simulation, it was possible to recreate a scanner resembling the Leica ScanStation P40. The scanner specifications were taken from the datasheet~\footnote{\url{https://leica-geosystems.com}} and implemented within Helios++ and are listed in Table \ref{tab:scanner}. To achieve a uniform sampling density, the vertical and horizontal angular resolutions were set to the same value of \SI{0.174}{\degree}. Accordingly, the head rotation speed was set to \SI{13.05}{\meter\per\second}, the pulse frequency to \SI{125000}{\hertz}, and the scan frequency to \SI{75}{\hertz}.

The setup of the simulation environments includes precise placement, translation, and scaling of the 3D objects, as well as careful positioning of the laser scanner around, and when applicable within, the objects. For larger-scale datasets, a solution was required to make the setup process faster, more intuitive, and supported by a visual interface for verification. To address this, a Blender\footnote{\url{https://www.blender.org}} add-on was developed that allows users to configure 3D scanning environments for Helios++ directly within Blender’s viewport. Using the add-on, scanners can be positioned accurately at suitable locations, and scanner settings can be adjusted while the resulting angular resolutions are displayed immediately. When a simulation is executed, the add-on automatically generates the scene and survey files in Extensible Markup Language (XML) format, which serve as the required inputs for Helios++. After the simulation is completed, the point clouds from each scanner are converted to ASTM E57 (E57) format, with the scanner location embedded in the files. These point clouds are unified into a single E57 file containing the XYZ coordinates and scanner positions, while individual scans remain distinguishable within the file. Finally, all output files are organised according to the PC2Model directory structure. The add-on is maintained on GitHub, which also provides the files used to calculate the dataset statistics: \href{https://github.com/saidharb/PC2Model.git}{https://github.com/saidharb/PC2Model.git}




\section{Evaluation strategy}\label{Evaluation}
To establish a consistent and rigorous evaluation strategy, we apply a known rigid body transformation to each CAD model. The parameters of this transformation serve as the ground truth targets for algorithms performing point cloud-to-model co-registration. We will use ICP as the baseline benchmark for our dataset for researchers to compare their results. For evaluation, we employ established metrics commonly used in related literature to assess both geometric alignment quality and spatial completeness. In addition, we evaluate the accuracy of the estimated rigid transformation to quantify co-registration performance.

\paragraph{Level of Accuracy (LOA):}
The geometric accuracy of the registration could be measured using the Level of Accuracy (LOA), which quantifies how closely the registered point cloud aligns with the reconstructed 3D model surface~\cite{ersoz_2026}. We employ two complementary distance-based measures to capture both overall alignment quality and robustness to outliers.

Let $\mathcal{P} = \{\mathbf{p}_i \in \mathbb{R}^3 \mid i = 1,\dots,N\}$ denote the registered point cloud and $\mathcal{S}$ represent the surface of the 3D model. For each point $\mathbf{p}_i$, the closest point $\mathbf{q}_i \in \mathcal{S}$ on the model surface by orthogonal projection is determined. The mean closest-point unsigned distance measures the average geometric deviation between the point cloud and the surface.

\begin{equation}
\mathrm{LOA}_{\mathrm{mean}} =
\frac{1}{N} \sum_{i=1}^{N}
\left\lVert \mathbf{p}_i - \mathbf{q}_i \right\rVert_2 
\end{equation}

In addition to the mean distance, we compute the median signed orthogonal distance between the point cloud and the surface. Let $d_i$ denote the signed distance from point $\mathbf{p}_i$ to the surface $\mathcal{S}$, where the sign is defined with respect to the local surface normal. This metric is more robust to outliers and allows the detection of systematic alignment biases.

\begin{equation}
\mathrm{LOA}_{\mathrm{median}} =
\mathrm{median}_{i=1,\dots,N} \left( d_i \right)
\end{equation}

\paragraph{Level of Coverage (LOC):}
While LOA focuses on geometric accuracy, it does not capture whether the entire model surface is sufficiently represented by the point cloud. To assess spatial completeness, we employ the Level of Coverage (LOC)~\cite{ersoz_2026}.

A regular grid $\mathcal{G} = \{\mathbf{g}_j \in \mathcal{S} \mid j = 1,\dots,M\}$ is sampled on the model surface. A grid point $\mathbf{g}_j$ is considered covered if there exists at least one point $\mathbf{p}_i \in \mathcal{P}$ whose Euclidean distance to $\mathbf{g}_j$ is below a predefined threshold $\tau$.

\begin{equation}
\left\lVert \mathbf{p}_i - \mathbf{g}_j \right\rVert_2 \leq \tau 
\end{equation}

The Level of Coverage is then defined as the ratio of covered grid points to the total number of sampled grid points, using the indicator function $\mathbb{I}(\cdot)$.

\begin{equation}
\mathrm{LOC} =
\frac{1}{M}
\sum_{j=1}^{M}
\mathbb{I}
\left(
\min_{\mathbf{p}_i \in \mathcal{P}}
\left\lVert \mathbf{p}_i - \mathbf{g}_j \right\rVert_2
\leq \tau
\right)
\end{equation}

\begin{figure}[h]
    \centering
    \includegraphics[width=\linewidth]{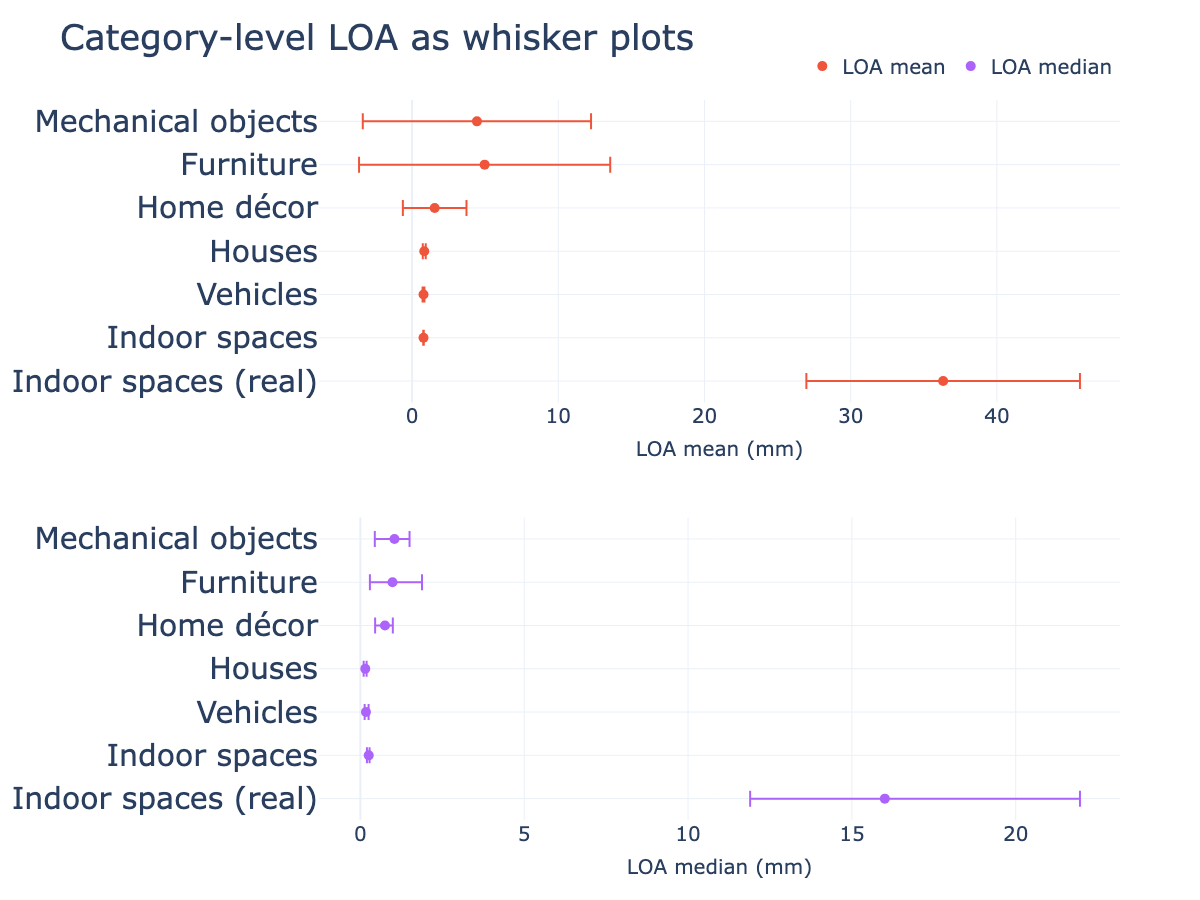}
    \caption{Category-level LOA of PC2Model objects.}
    \label{fig:loa_stats}
\end{figure}

\begin{figure}[bh]
    \centering
    \includegraphics[width=\linewidth]{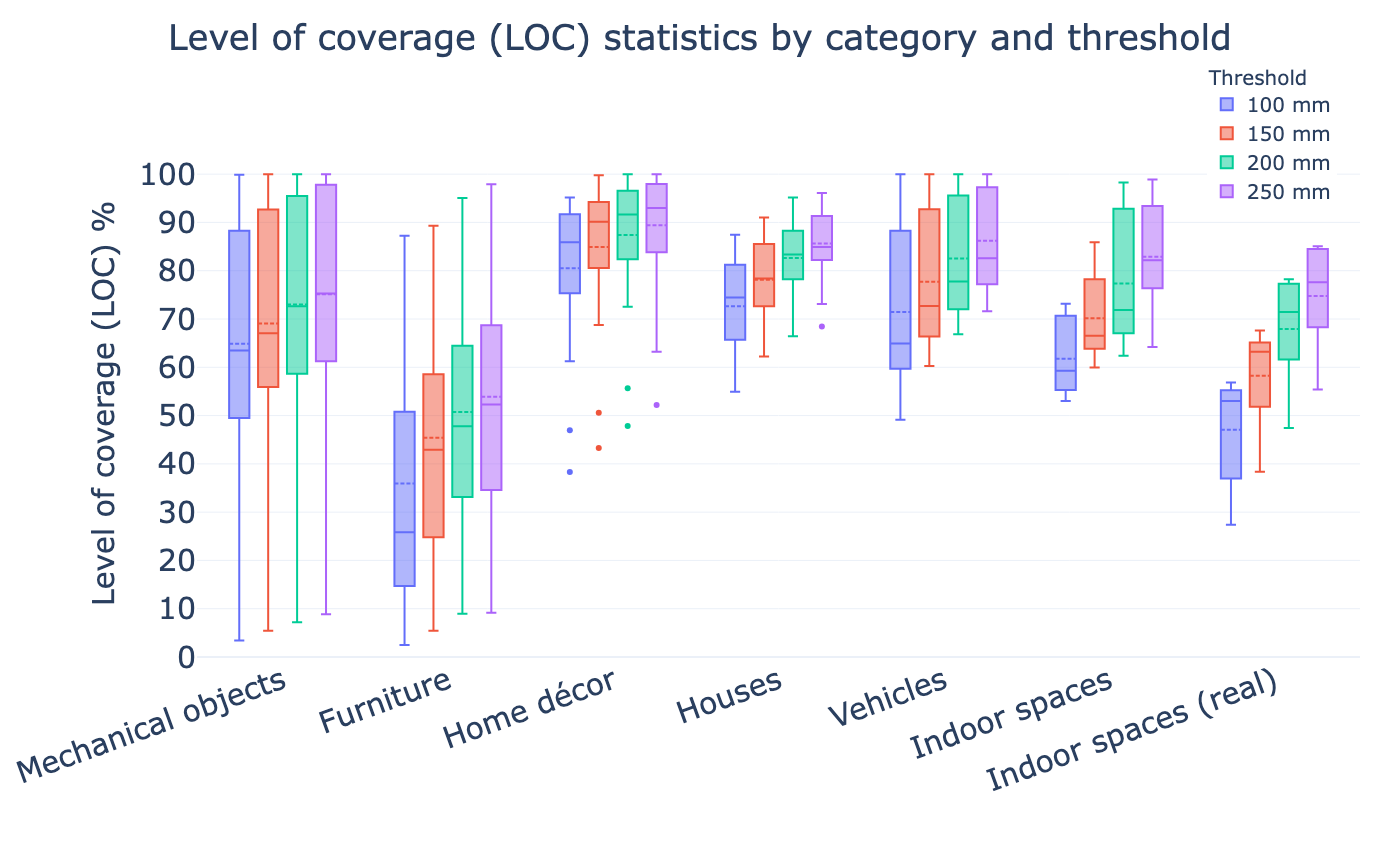}
    \caption{Level of Coverage (LOC) distributions across categories for different thresholds.}
    \label{fig:loc_stats}
\end{figure}

In order to provide deeper insights into the geometric quality and spatial completeness of the benchmark data for point-to-mesh registration across different object categories, the category-level distributions of LOA and LOC are visualised in Figures~\ref{fig:loa_stats} and~\ref{fig:loc_stats}, respectively. It is worth mentioning that the same metrics could be used for the evaluation of point cloud to model co-registration algorithms.

As it is visible in Figure~\ref{fig:loa_stats}, all simulated categories (e.g., vehicles, houses, indoor, and home décor) achieve mean LOA values of better than 1 cm, demonstrating high geometric alignment accuracy. As expected indoor spaces (real) exhibit higher LOA values reflecting clutter, and higher registration error in real-world data. 

Figure~\ref{fig:loc_stats} presents the LOC distributions across categories for multiple distance thresholds (10 cm to 25 cm). As expected, LOC increases with higher threshold values for all categories. Simulated categories generally achieve high coverage, indicating that most of the model's surfaces are well represented by the point cloud. However, categories like furniture show lower coverage at stricter thresholds, mainly due to occlusions. Indoor spaces (real) demonstrate comparatively lower LOC at smaller thresholds but improve significantly at larger thresholds.

\paragraph{Transformation Error Metrics:}

The accuracy of the estimated transformation is evaluated by comparing it 
to the ground truth using the error matrix $\mathbf{T}_{\mathrm{err}} \in \mathrm{SE}(3)$, 
defined as:

\begin{equation}
    \mathbf{T}_{\mathrm{err}} = \mathbf{T}_{\mathrm{est}}^{-1} \mathbf{T}_{\mathrm{gt}}=
\begin{bmatrix}
\mathbf{R}_{\mathrm{err}} & \mathbf{t}_{\mathrm{err}} \\
\mathbf{0}^\top & 1
\end{bmatrix}
\end{equation}
where $\mathbf{T}_{\mathrm{gt}}$ is the ground truth transformation from point cloud to model  and 
$\mathbf{T}_{\mathrm{est}}$ is the estimated transformation from point cloud to model. A perfect estimate 
yields $\mathbf{T}_{\mathrm{err}} = \mathbf{I}$.

The \textit{translation error} ($e_t$) is computed as the Euclidean norm of the translation vector. To make translation errors comparable across different point densities, they are divided by the length of their bounding box diagonal $d$:
\begin{equation}
    e_t = \frac{\lVert \mathbf{t}_{\mathrm{err}} \rVert_2}{d}
\end{equation}

The \textit{rotation error} ($e_r$) measures the difference between the estimated rotation and the ground truth rotation~\cite{Hartley_Zisserman_2004}. 
Formally, it is defined as the \emph{geodesic distance} on the rotation manifold SO(3), 
which corresponds to the minimal angle of rotation needed to align the two rotations.

Let $\mathbf{R}_{\mathrm{err}}$ denote the relative rotation between the estimated and ground truth rotations:

\begin{equation}
    \mathbf{R}_{\mathrm{err}} = \mathbf{R}_{\mathrm{est}}^{-1} \mathbf{R}_{\mathrm{gt}}
\end{equation}

The rotation error $e_r$ is then computed from the rotation matrix $\mathbf{R}_{\mathrm{err}}$ as:

\[
e_r = \arccos\left( \frac{\mathrm{trace}(\mathbf{R}_{\mathrm{err}}) - 1}{2} \right)
\]

where $\mathrm{trace}(\cdot)$ denotes the sum of the diagonal elements of a matrix.

\textbf{Baseline evaluation:} We use ICP to quantify the achievable alignment accuracy of a standard local registration method under controlled initial conditions. ICP is a local optimisation method and therefore requires a reasonably good initial alignment (rough co-registration). To obtain a meaningful baseline, we applied only a moderate transformation to the point clouds: translations were limited to at most $0.1$ times the maximum bounding-box extent per axis, rotations were restricted to the range $[-10^\circ, 10^\circ]$, and no scaling was applied, in contrast to the much larger transformations in our final dataset. For ICP, a uniformly sampled point cloud of the 3D model serves as the reference. ICP is then performed to align the moderately transformed scan back to this reference. The ICP algorithm was applied using Open3D~\cite{o3d} and CloudCompare\footnote{\url{https://www.cloudcompare.org}} with a point-to-plane error metric and a maximum of 2000 iterations. The convergence criterion was set to an RMS difference of $1.0 \times 10^{-5}$. To control computational cost and avoid bias from extremely dense scans, the transformed point cloud was randomly downsampled to at most 50{,}000 points prior to registration, preserving the global geometric structure while significantly reducing runtime and memory requirements. The results can be seen in Table~\ref{tab:category_metrics}.

\begin{table}[h]
\centering
\setlength{\tabcolsep}{4pt}
\resizebox{\columnwidth}{!}{%
\begin{tabular}{l l S[table-format=2.3] c}
\toprule
Source & Category & {Avg. $e_t$ in mm} & Avg. $e_r$ in \(^\circ\) \\
\midrule
\multicolumn{1}{c|}{\multirow{6}{*}{\makecell[c]{Simulated\\from\\3D model}}}
    & Mechanical objects & 4 & 1.2 \\
\multicolumn{1}{c|}{}
    & Furniture          & 1 & 0.0 \\
\multicolumn{1}{c|}{}
    & Home décor         & 2 & 0.5 \\
\multicolumn{1}{c|}{}
    & Houses             & 0 & 0.0 \\
\multicolumn{1}{c|}{}
    & Vehicles           & 0 & 0.0 \\
\multicolumn{1}{c|}{}
    & Indoor spaces      & 0 & 0.0 \\
\cline{1-1}
\multicolumn{1}{c|}{\raisebox{-2pt}{Real}}
    & {\raisebox{-2pt}{Indoor spaces}}      & 5 & 1.0 \\
\bottomrule
\end{tabular}%
}
\caption{ICP alignment errors under coarse initialization: normalized translation error $e_t$ and rotation error $e_r$.}
\label{tab:category_metrics}
\end{table}

The PC2Model dataset poses a range of algorithmic challenges for point cloud registration methods. First, the dataset contains a large variety of object categories and corresponding shapes, which prevents algorithms from being trained on a single object class and instead requires strong generalisation capabilities. This diversity extends beyond object categories to the properties of the point clouds themselves: object sizes vary significantly, point densities differ across scans, and different scanner positions result in varying and incomplete views of the same object. Due to the simulation capabilities of Helios++, the generated point clouds exhibit a high degree of realism. Rather than idealised scans, the data contain real-world artefacts such as occlusions, partial scans, and point accuracy deviations. Point outliers, however, are a comparatively minor challenge, as objects are placed in empty scenes, making spurious measurements rare. Another major challenge lies in estimating the rigid transformation between point clouds. The transformations are randomly performed, leading to potentially large translations and rotations, such that initial misalignment can be substantial, although partial overlap may still be present. In addition, the dataset introduces significant computational challenges, as individual point clouds can contain up to 132 million points. Finally, even objects within the same category exhibit notable geometric variation, requiring registration algorithms to robustly account for diverse local and global features.

\section{Conclusions}\label{Conclusion}
The PC2Model benchmark dataset provides a controlled, scalable, and cost-effective resource comprising point clouds and corresponding 3D models for training and evaluating both conventional and deep learning-based registration methods. By combining simulated point clouds with real-world scans, the dataset overcomes limitations commonly associated with exclusively real-world data, such as noise, occlusions, and limited availability. This benchmark is intended to support the research community in developing, comparing, and validating registration methods, contributing to more robust and accurate solutions in the field.The dataset is publicly accessible at:\newline \href{https://doi.org/10.5281/zenodo.17581812}{https://zenodo.org/records/17581812}.

Future work will focus on extending the PC2Model dataset to increase its diversity and complexity for registration tasks. Planned enhancements include expanding object samples and categories, incorporating partial occlusions, and simulating varying noise levels to better reflect real-world scanning conditions. In addition, we plan to develop a web-based evaluation platform to facilitate the benchmarking of registration methods and visualisation of performance metrics.

\textbf{Acknowledgments:}
This work is funded by the ISPRS Scientific Initiatives 2025. It is also supported by the Deutsche Forschungsgemeinschaft (DFG, German Research Foundation)– TRR 277/2 2024– Project number 414265976. 

{
	\begin{spacing}{1.17}
		\normalsize
		\bibliography{ISPRSguidelines_authors} 
	\end{spacing}
}

\end{document}